\documentclass[letterpaper]{article} 
\usepackage{aaai25}  
\usepackage{times}  
\usepackage{helvet}  
\usepackage{courier}  
\usepackage[hyphens]{url}  
\usepackage{graphicx} 
\usepackage{natbib}  
\usepackage{caption} 
\usepackage{algorithm}
\usepackage{algorithmic}
\usepackage{amsmath}
\usepackage{booktabs}\usepackage{booktabs}
\usepackage{newfloat}
\usepackage{listings}
\DeclareCaptionStyle{ruled}{labelfont=normalfont,labelsep=colon,strut=off} 
\floatstyle{ruled}
\newfloat{listing}{tb}{lst}{}
\floatname{listing}{Listing}
\title{Alignment-Free RGB-T Salient Object Detection: A Large-scale Dataset and Progressive Correlation Network}
\author{
	Kunpeng Wang\textsuperscript{\rm 1}, 
	Keke Chen\textsuperscript{\rm 1},
	Chenglong Li\textsuperscript{\rm 2},
	Zhengzheng Tu\textsuperscript{\rm 1}\thanks{Corresponding authors.},
	Bin Luo\textsuperscript{\rm 1}\footnotemark[1]
}
\affiliations{
	\textsuperscript{\rm 1}Key Laboratory of Intelligent Computing and Signal Processing of Ministry of Education, Anhui Provincial Key Laboratory of Multimodal Cognitive Computation, School of Computer Science and Technology, Anhui University, China\\
	\textsuperscript{\rm 2}Anhui Provincial Key Laboratory of Security Artificial Intelligence, School of Artificial Intelligence, Anhui University, China\\
	
	\{kp.wang, chenk1220, lcl1314\}@foxmail.com, zhengzhengahu@163.com, luobin@ahu.edu.cn
}

\usepackage{bibentry}
\usepackage{rotating}
\usepackage{multirow}
\usepackage{bigstrut}
\usepackage{colortbl}
\usepackage{pifont}

\begin{document}

\maketitle

\begin{abstract}
	Alignment-free RGB-Thermal (RGB-T) salient object detection (SOD) aims to achieve robust performance in complex scenes by directly leveraging the complementary information from unaligned visible-thermal image pairs, without requiring manual alignment. However, the labor-intensive process of collecting and annotating image pairs limits the scale of existing benchmarks, hindering the advancement of alignment-free RGB-T SOD. In this paper, we construct a large-scale and high-diversity unaligned RGB-T SOD dataset named UVT20K, comprising 20,000 image pairs, 407 scenes, and 1256 object categories. All samples are collected from real-world scenarios with various challenges, such as low illumination, image clutter, complex salient objects, and so on. To support the exploration for further research, each sample in UVT20K is annotated with a comprehensive set of ground truths, including saliency masks, scribbles, boundaries, and challenge attributes.
	In addition, we propose a Progressive Correlation Network (PCNet), which models inter- and intra-modal correlations on the basis of explicit alignment to achieve accurate predictions in unaligned image pairs. Extensive experiments conducted on unaligned and aligned datasets demonstrate the effectiveness of our method.
	\begin{links}
		\link{Code and Dataset}{https://github.com/Angknpng/PCNet}
	\end{links}
\end{abstract}

\section{Introduction}
Salient object detection (SOD) aims to identify and segment the most attractive regions in visual scenes. It can help eliminate redundant information and has been applied in a variety of vision tasks, such as image compression~\cite{li2017closed}, video analysis~\cite{fan2019video}, and visual tracking~\cite{zhang2020track}. Although great progress has been made, RGB SOD methods~\cite{wang2022survey} only based on visible images still struggle to distinguish salient regions in complex scenes, such as background clutter, low illumination, and similar foreground and background. 

Thermal sensors can capture the overall shape of objects, providing complementary information to visible images~\cite{lu2021rgbt,zhu2021rgbt,yuan2024improving}. Therefore, some studies introduce thermal images alongside visible images to enhance performance in complex scenes.
For example, \cite{tu2020rgbt} release a prevalent RGB-Thermal (RGB-T) SOD benchmark with spatial alignment to facilitate the utilization of corresponding multi-modal information. 
Therefore, existing methods are almost designed on the basis of alignment, and demonstrate their effectiveness through modality discrepancy reduction~\cite{liu2021swinnet}, multi-modal interaction improvement~\cite{tu2021multi}, modality modulation~\cite{cong2022does}, and so on. 

\begin{figure}[t]
	\centering
	\includegraphics[width=1\columnwidth]{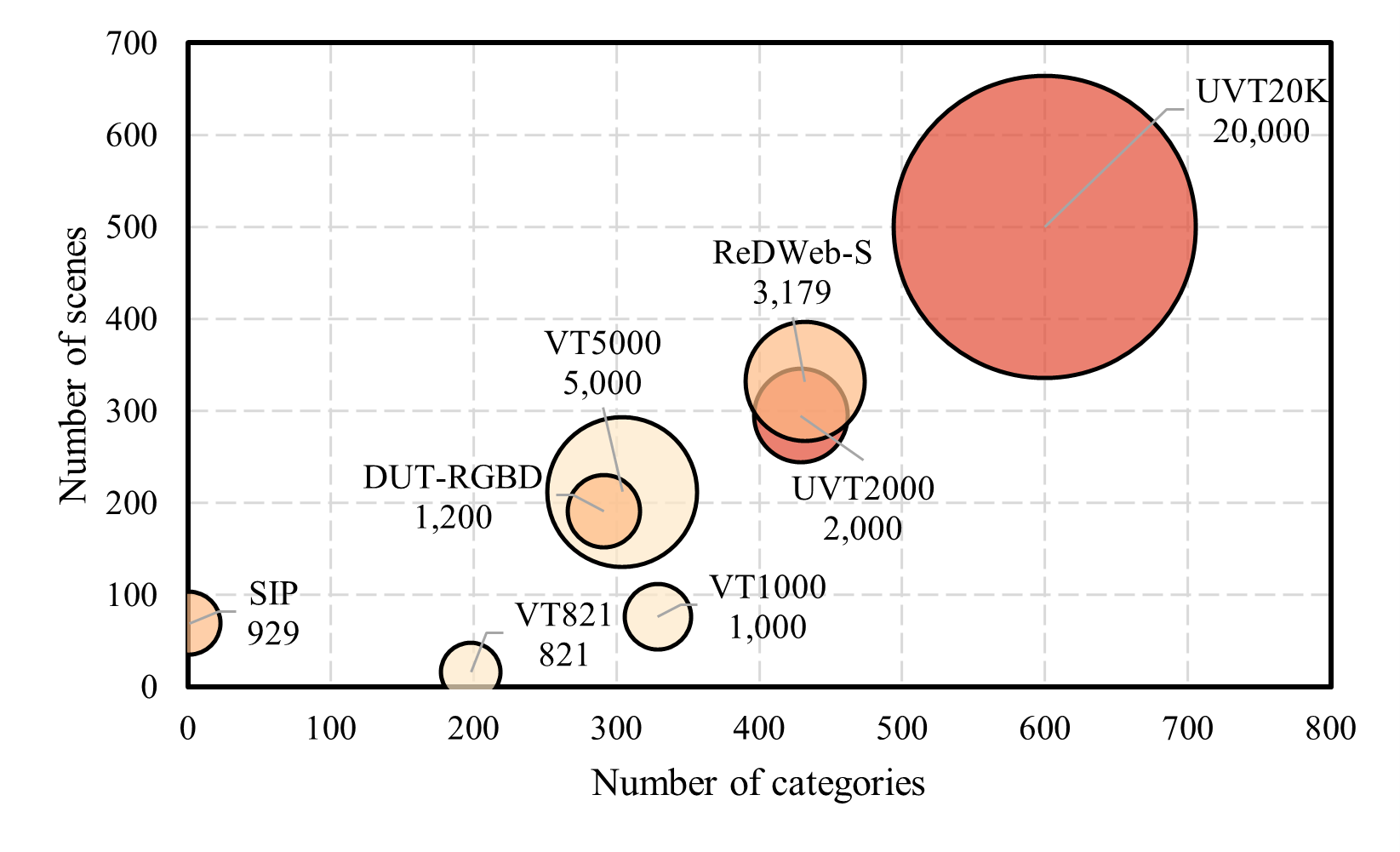}
	\caption{Comparison on scale (i.e., circular area), scenes (i.e., vertical axis), and object categories (i.e., horizontal axis) of the proposed UVT20K dataset with existing representative RGB-T and RGB-D SOD datasets, including UVT2000~\cite{wang2024alignment}, VT5000~\cite{tu2020rgbt}, VT1000~\cite{tu2019rgb}, VT821~\cite{wang2018rgb}, ReDWeb-S~\cite{liu2021learning}, SIP~\cite{fan2020rethinking}, and DUT-RGBD~\cite{piao2019depth}.}
	\label{fig::dataset_com}
\end{figure}
However, the original captured RGB-T image pairs are unaligned in space and scale, and manually aligning them is labor-intensive and not conducive to practical application and deployment. Moreover, directly applying existing alignment-based methods to unaligned data leads to significant performance degradation due to the difficulty of exploiting multi-modal correspondences. To this end, \cite{tu2022weakly} make the first attempt to build modality correspondences on weakly aligned datasets, which are artificially created from existing aligned datasets~\cite{wang2018rgb,tu2019rgb,tu2020rgbt} by spatial affine transformation.
While the weakly aligned datasets contribute positively to the advancement of SOD, they still fall short of meeting practical application requirements. In addition, the local correspondences established through convolutional operations are insufficient to handle large spatial deviations that occur in real-world misalignment.
Recently, \cite{wang2024alignment} release the first unaligned dataset (i.e., UVT2000) and design a correlation modeling method based on asymmetric window pairs. Nonetheless, two issues still exist: 1) UVT2000 is limited in scale as it only contains 2,000 unaligned image pairs for testing and lacks a training set for models to learn the properties of unaligned data, 2) the fixed-size asymmetric window pairs cannot flexibly adjust their shapes according to the misalignment status in different scenes, failing to cover the complete corresponding multi-modal information.

To address the first issue, we construct a large-scale unaligned RGB-T SOD benchmark named UVT20K with multiple characteristics. Firstly, to the best of our knowledge, UVT20K is the largest multi-modal SOD dataset containing 20,000 unaligned image pairs with both training and testing sets. It will play an important role in training and evaluating alignment-free RGB-T SOD methods. Secondly, UVT20K covers 407 scenes, 1256 object categories, and 15 kinds of challenges, endowing it with a high degree of diversity and complexity. Thirdly, each sample in UVT20K is annotated with a comprehensive set of ground truths, including saliency masks, scribbles, boundaries, and challenge attributes, which provide a foundation for extensive research.
As shown in Fig.~\ref{fig::dataset_com}, UVT20K is far beyond recent representative multi-modal datasets in terms of scale, scene, and object category. Notably, we include both RGB-T and RGB-Depth (RGB-D) datasets for a comprehensive comparison.

To address the second issue, we propose an Progressive Correlation Network (PCNet) that models inter- and intra-modal correlations on the basis of explicit alignment. To be specific, we propose a Semantics-guided Homography Estimation (SHE) module that introduces and fine-tunes an existing multi-modal homography estimator~\cite{cao2022iterative} to explicitly align the common regions between RGB and thermal modalities. Considering that the estimator is pre-trained on other multi-modal datasets, we design an S-Adapter to adapt it to RGB-T data. Meanwhile, semantic information is embedded into S-Adapter to guide the estimator to predict the homography matrix against object regions.
Since only the overlapping multi-modal regions will be aligned in unaligned image pairs, it is difficult to model the correlation of complete saliency regions with multi-modal fusion alone.
To this end, we also propose an Inter- and Intra-Modal Correlation (IIMC) module to fully model the correlation of salient regions. 
IIMC first models the multi-modal correlation of the corresponding partial region, and then expands the correlation to the entire region within the modality. In this way, correlations of salient objects can be modeled progressively to achieve accurate saliency predictions. The main contributions of our work are as follows:
\begin{itemize}
	\item We construct a large-scale benchmark dataset named UVT20K for alignment-free RGB-T SOD. To the best of our knowledge,  UVT20K is the largest multi-modal SOD dataset, containing 20,000 unaligned visible-thermal image pairs, covering 407 scenes and 1256 object categories with 15 challenges, providing a comprehensive set of annotations. Our dataset will be available to the research community for extensive investigation and exploration.
	\item We propose a Progressive Correlation Network (PCNet) that explores and integrates saliency cues in unaligned RGB-T image pairs to achieve accurate predictions.
	\item We propose a semantics-guided homography estimation module to explicitly align corresponding multi-modal regions, and an inter- and intra-modal correlation module to progressively model correlations for salient objects.
	\item We conduct a comprehensive evaluation of state-of-the-art methods and an in-depth experimental analysis of our PCNet on the newly constructed benchmark dataset, as well as the other aligned and unaligned datasets.
\end{itemize}

\section{Related Work}
\subsection{RGB-T Salient Object Detection Benchmarks}
The first dataset for RGB-T Salient Object Detection (SOD) is VT821~\cite{wang2018rgb}, which contains 821 pairs of images collected from a recording system consisting of a thermal imager (i.e., FLIR A310) and a CCD camera (i.e., SONY TD-2073). All 821 image pairs are manually aligned using point correspondences and annotated with saliency masks and challenge attributes. Then, VT1000~\cite{tu2019rgb} enlarges the scale to 1,000 manually aligned and annotated image pairs, which are captured by the FLIR SC620 device with a thermal infrared camera and a CCD camera inside. Later, VT5000~\cite{tu2020rgbt} was built with 5,000 image pairs captured by the FLIR T640 device. The aligned samples in VT5000 are divided into two halves, each containing 2,500 samples for the training and testing. These aligned datasets provide the foundation for alignment-based methods, however, the labor-intensive manual alignment is not conducive to practical application and deployment. 

To overcome this problem, ~\cite{tu2022weakly} artificially create the corresponding weakly aligned datasets (i.e., un-VT821, un-VT1000, and un-VT5000) by performing random affine transformations on the above three aligned datasets. Recently, \cite{wang2024alignment} release the UVT2000 dataset with 2,000 unaligned image pairs captured by a FLIR SC620 device without manual alignment. However, UVT2000 is small in scale and lacks a training set, preventing the corresponding methods from learning the properties of unaligned image pairs and limiting the research on alignment-free RGB-T SOD. To this end, we construct a large-scale unaligned dataset consisting of 20,000 image pairs, which contain a training set and a test set, rich scenes and object categories, and a variety of annotations.

\begin{figure*}[t]
	\centering
	\includegraphics[width=1\linewidth]{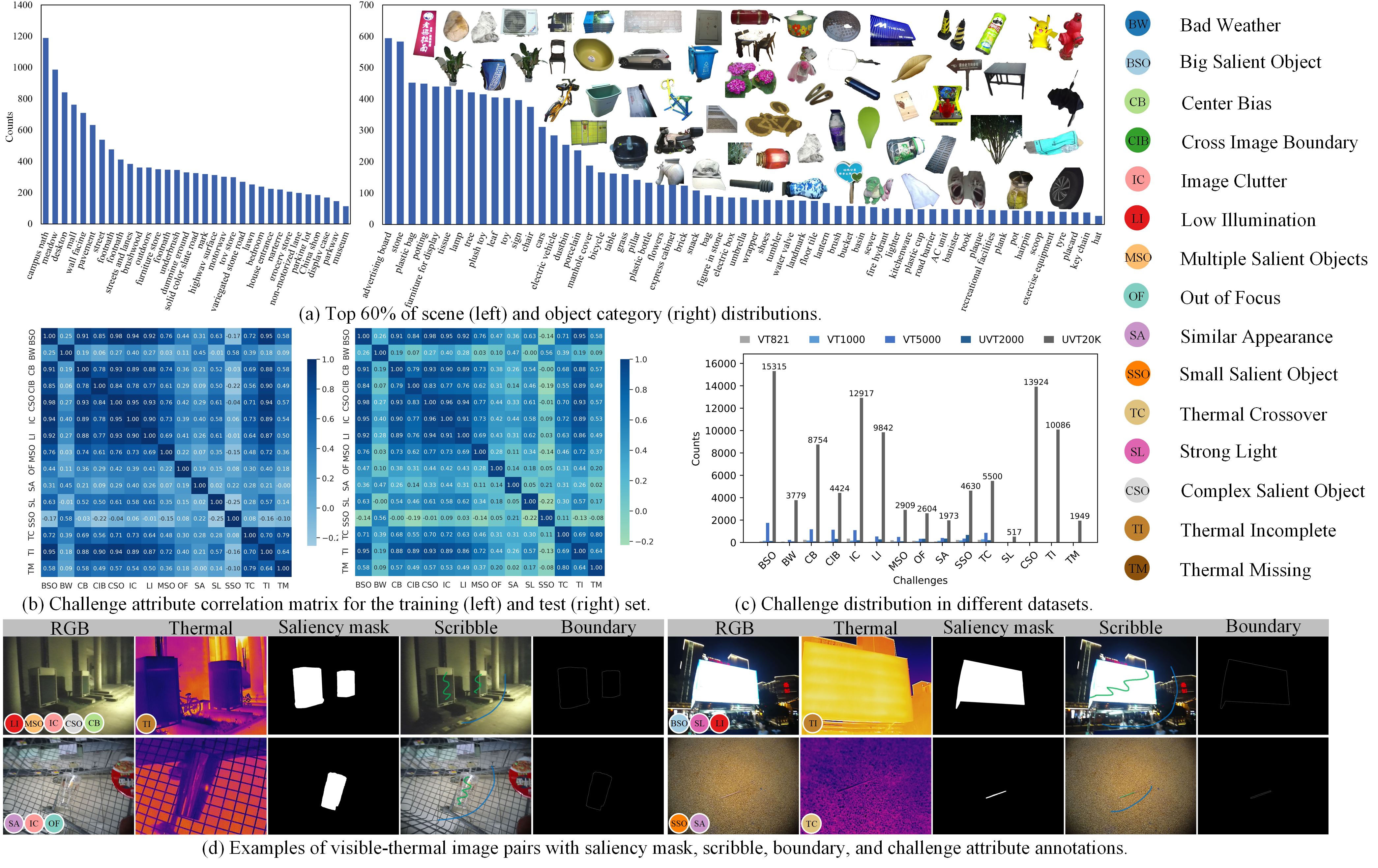}
	\caption{Main statistics and characteristics of our UVT20K dataset.} 
	\label{fig::daatset_property}
\end{figure*}
\subsection{RGB-T Salient Object Detection Methods}
RGB-T salient object detection focuses on fusing information from RGB and thermal modalities to predict their common salient regions. Based on the aligned benchmark datasets, enormous methods have been presented through graph learning~\cite{tang2019rgbt,tu2019rgb}, multi-scale feature interaction~\cite{wang2021cgfnet,tu2021multi}, lightweight design~\cite{huo2021efficient,huo2022real,zhou2023lsnet}, global information modeling~\cite{liu2021swinnet,tang2022hrtransnet,pang2023caver}, and ensemble learning~\cite{wang2024learning}. Although great progress has been made, these methods rely on the corresponding spatial information in modality alignment. Therefore, it is difficult for them to capture the correlated information in unaligned image pairs. 

DCNet~\cite{tu2022weakly} attempts to establish multi-modal correspondences in weakly aligned image pairs via affine transformation and dynamic convolution. Although effective in weakly aligned image pairs with small spatial offsets, the local receptive fields of the convolutional operation have difficulty in handling large deviations in space and scale. To improve it, SACNet~\cite{wang2024alignment} uses a pair of asymmetric windows to cover the corresponding information in unaligned image pairs. However, the fixed windows fail to flexibly cover the corresponding information for different scenes, leading to the introduction of unrelated noise for correlation modeling. In this paper, we explicitly align the common information of the two modalities and progressively model inter- and intra-modal correlations.

\begin{table*}[t]
	\centering
	\resizebox{1\textwidth}{!}{
		\begin{tabular}{c|c|ccc|c|c|c|c|c|c|c|c|c|c}
			\toprule[1pt]
			\multirow{2}[4]{*}{Dataset} & \multirow{2}[4]{*}{Year} & \multicolumn{3}{c|}{Modality} & Scene & Category & Alignment & Practical & Training & Weak  & Boundary & Challenge & Challenge & \multirow{2}[4]{*}{Image Source} \bigstrut\\
			\cline{3-5}          &       & RGB   & Thermal & Depth & Number & Number & Status & Shooting & Subset & Annotation & Annotation & Annotation & Number &  \bigstrut\\
			\hline
			DUTLF-Depth & 2019  & 1200  & -     & 1200  & 191   & 291   & Aligned & \ding{51} & \ding{51} & \ding{53}     & \ding{53}     & \ding{53}     & -     & By Lytro camera \bigstrut[t]\\
			SIP   & 2020  & 929   & -     & 929   & 69    & 1     & Aligned & \ding{51} & \ding{53}     & \ding{53}     & \ding{53}     & \ding{53}     & -     & By mobile phone \\
			ReDWeb-S & 2021  & 3,179 & -     & 3,179 & 332   & 432   & Aligned & \ding{53}     & \ding{51} & \ding{53}     & \ding{53}     & \ding{53}     & -     & From ReDWeb dataset \bigstrut[b]\\
			\hline
			VT821 & 2018  & 821   & 821   & -     & 16    & 198   & Aligned & \ding{51} & \ding{53}     & \ding{53}     & \ding{53}     & \ding{51} & 11    & By FLIR camera \bigstrut[t]\\
			VT1000 & 2019  & 1,000 & 1,000 & -     & 76    & 329   & Aligned & \ding{51} & \ding{53}     & \ding{53}     & \ding{53}     & \ding{51} & 10    & By FLIR camera \\
			VT5000 & 2020  & 5,000 & 5,000 & -     & 212   & 304   & Aligned & \ding{51} & \ding{51} & \ding{53}     & \ding{53}     & \ding{51} & 11    & By FLIR camera \\
			un-VT821 & 2022  & 821   & 821   & -     & 16    & 198   & Weakly aligned & \ding{53}     & \ding{53}     & \ding{53}     & \ding{53}     & \ding{53}     & -     & From VT821 dataset \\
			un-VT1000 & 2022  & 1,000 & 1,000 & -     & 76    & 329   & Weakly aligned & \ding{53}     & \ding{53}     & \ding{53}     & \ding{53}     & \ding{53}     & -     & From VT1000 dataset \\
			un-VT5000 & 2022  & 5,000 & 5,000 & -     & 212   & 304   & Weakly aligned & \ding{53}     & \ding{51} & \ding{53}     & \ding{53}     & \ding{53}     & -     & From VT5000 dataset \\
			UVT2000 & 2024  & 2,000 & 2,000 & -     & 295   & 429   & Unaligned & \ding{51} & \ding{53}     & \ding{53}     & \ding{53}     & \ding{51} & 11    & By FLIR camera \bigstrut[b]\\
			\hline
			UVT20K (Ours) & -     & \textbf{20,000} & \textbf{20,000} & -     & \textbf{407} & \textbf{1256} & Unaligned & \ding{51} & \ding{51} & \ding{51} & \ding{51} & \ding{51} & \textbf{15} & By Hikvision and FLIR cameras \bigstrut\\
			\bottomrule[1.2pt]
		\end{tabular}%
	}
	\caption{Comparison of UVT20K with prevalent multi-modal SOD datasets.}
	\label{tab:dataset}%
\end{table*}%
\section{UVT20K Benchmark}

Existing RGB-T SOD datasets~\cite{wang2018rgb,tu2019rgb,tu2020rgbt} are almost manually aligned and labor-intensive, while the only unaligned dataset~\cite{wang2024alignment} is limited in scale and lacks a training set, which restricts the research on alignment-free RGB-T SOD. To address this issue, we construct a large-scale and high-diversity unaligned RGB-T SOD dataset denoted as UVT20K.

\subsection{Dataset Construction}
\noindent\textbf{Dataset Collection.}
The proposed UVT20K dataset is captured in the real world using Hikvision DS-2TP23-10VF/W(B) and FLIR SC620 devices, which are equipped with a pair of CCD and thermal infrared cameras. 
Each sample is directly obtained by camera shot without manual alignment. Although integrated in a single device, the CCD and thermal infrared camera pairs have parallax and different viewing angle sizes. Therefore, the same object captured by both devices suffers from positional offsets and scale differences, resulting in misalignment.

\noindent\textbf{Dataset Annotation.}
We first perform a quality scan on the initially captured 24,000+ image pairs to remove duplicate, objectless, and corrupted samples. 
Then, the rest 22,000+ high-quality image pairs are delivered to ten annotators, who select and annotate the most attractive objects or regions at the first glance. Based on the majority voting for salient objects, we rank and select the top 20,000 annotated samples as the final dataset. 
Note that the annotation process is based on the human-preferred RGB images, with the corresponding thermal images used for assistance.
In addition, we annotate each sample with scribbles and boundaries, which can be used for weak supervision and other related studies. To facilitate subsequent work on challenging scenes, we also annotate each sample with one or more challenge attributes. A total of 15 challenge attributes are listed on the right side of Fig.\ref{fig::daatset_property}, in which the first 11 challenge attributes are derived from existing RGB-T datasets~\cite{tu2020rgbt} and the last four new challenges (i.e., SL, CSO, TI, TC) are summarized and added for UVT20K. In particular, TI is the case where salient objects are not completely captured in thermal images, and TC is the extreme case of TI. Some visual examples from our dataset are illustrated in Fig.~\ref{fig::daatset_property}(d). Due to space limitations, more examples are provided in $supp$.

\noindent\textbf{Dataset Splits.}
Following the representative RGB-T SOD dataset VT5000~\cite{tu2020rgbt}, we randomly select 10,000 pairs of images in UVT20K dataset as the training set, and take the remaining 10,000 pairs as the test set. Fig.~\ref{fig::daatset_property}(b) illustrates the challenge attribute correlation matrix for the training and test sets. A higher correlation between two challenge attributes in the matrix indicates a higher probability of their co-existence. The similar challenge distributions validate the rationality of the dataset splits and facilitate the accurate evaluation of various models.

\subsection{Dataset Comparisons and Characteristics}
As shown in Table~\ref{tab:dataset}, compared with prevalent multi-modal SOD datasets, UVT20K has the following characteristics.
\begin{itemize}
	\item UVT20K contains a vast collection of 20,000 unaligned image pairs with both training and test sets. To the best of our knowledge, it is the largest multi-modal SOD dataset, which will advance research on alignment-free RGB-T SOD, saving the labor cost of manual alignment.
	\item UVT20K covers a wide range of scenes and object categories. As in previous work~\cite{liu2021learning,wang2024alignment}, we count and report the total number of scenes and object categories for UVT20K and the compared datasets in Table~\ref{tab:dataset}, which suggests that UVT20K has the most variety of scenes and objects. For a more detailed analysis, Fig.~\ref{fig::daatset_property}(a) illustrates the counts of the top 60\% scenes and object categories, showing an approximately smooth distribution.  
	\item UVT20K provides a more comprehensive set of annotations, including fully-supervised saliency masks, weakly-supervised scribbles, object boundaries, and challenge attributes. Compared to existing multi-modal SOD datasets that typically involve only saliency mask annotations, the diverse annotation sets in UVT20K enable the exploration for further research, such as weakly-supervised methods and boundary-enhanced algorithms.
	\item UVT20K presents significant challenges for exploration. As shown in  Table~\ref{tab:dataset}, UVT20K has the highest number of challenge types. Fig.~\ref{fig::daatset_property}(c) shows that UVT20K has the largest count within each challenge category, and Fig.~\ref{fig::daatset_property}(b) indicates that each UVT20K sample typically encompasses multiple challenges. In addition, multi-platform shoots at Hikvision and FLIR further intensify the challenges. These challenges will be a valuable asset for assessing the robustness of SOD models.
\end{itemize}

\section{Methodology}
\begin{figure*}[t]
	\centering
	\includegraphics[width=1\linewidth]{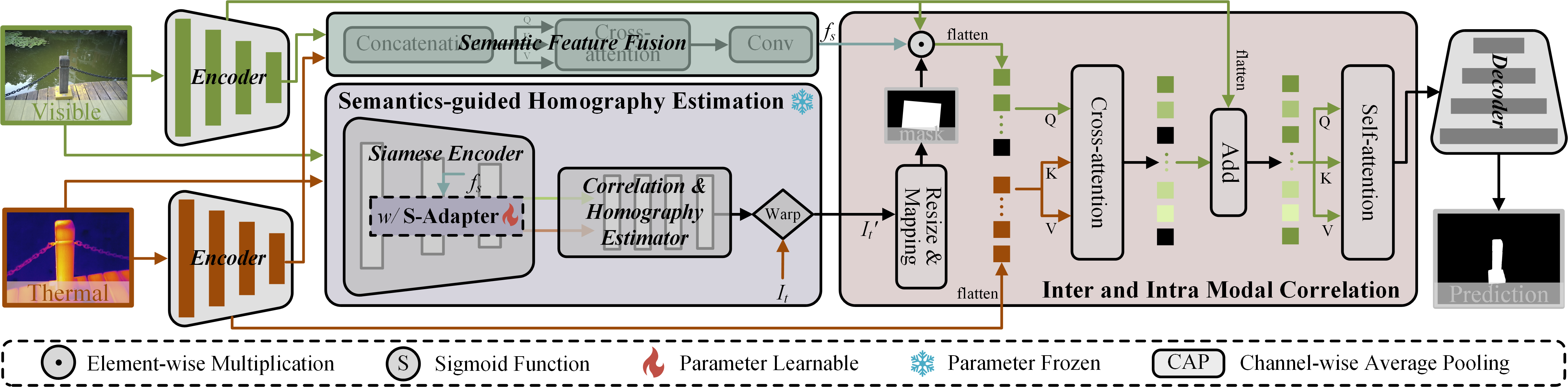}
	\caption{The overall architecture of our proposed Progressive Correlation Network (PCNet). The framework mainly comprises a Semantics-guided Homography Estimation (SHE) module and an Inter- and Intra-Modal Correlation (IIMC) module. SHE is fine-tuned by the S-Adapter to explicitly align the corresponding regions in visible-thermal image pairs. IIMC first models inter-modal correlations for the aligned regions, and then expand the correlations to the whole RGB modality.} 
	\label{fig::framework}
\end{figure*}
\noindent\textbf{Motivation.}
For aligned visible-thermal image pairs, the same object is corresponding in space. In this case, the RGB and thermal modalities are inherently correlated, which provides a basis for further exploration of multi-modal complementary information.
However, there are positional offsets and scale differences between unaligned image pairs, reducing the modality correlation.
Therefore, modeling the corresponding multi-modal correlations is the key to alignment-free RGB-T SOD. 
To this end, the proposed Progressive Correlation Network (PCNet) first seeks the corresponding regions in unaligned image pairs, and then models the inter- and intra-modal correlations for salient regions.

\subsection{Overview}
The overall architecture of PCNet is illustrated in Fig.~\ref{fig::framework}, which receives a pair of unaligned RGB and thermal images and consists of two parallel encoders, a semantic feature fusion component, a Semantics-guided Homography Estimation (SHE) module, an Inter- and Intra-Modal Correlation (IIMC) module, and a decoder. 
To be specific, the two encoders (i.e., Swin-B~\cite{liu2021swin}) separately extract multi-level features from the input image pairs (i.e., $I_{rgb}$ and $I_{t}$), denoted as $f_{m}^i$ ($m \in \left\{ {rgb,t} \right\},i = 1,...,4$). 
The semantic feature fusion component integrates the top extracted features (i.e., $f_{rgb}^4$ and $f_{t}^4$) through attention operation to obtain semantic information (i.e., ${f_s}$) embedded in SHE and IIMC. 
SHE is proposed to predict the corresponding regions between RGB and thermal image pairs based on homography estimation.
On this basis, IIMC models inter- and intra-modal correlations specific to salient regions.
Then, the decoder integrates the correlated multi-modal and multi-level features to obtain the final prediction.

\subsection{Semantics-guided Homography Estimation (SHE)}
The same object in unaligned RGB and thermal image pairs is inconsistent in position and scale, resulting in only partial regions correspond to each other. In this case, directly modeling the correlation between the two modalities introduces too much irrelevant information causing noise interference. In order to find the corresponding regions in the RGB and thermal image pairs, we introduce a pre-trained multi-modal homography estimator IHN~\cite{cao2022iterative}, which is structurally simple but effective. 
However, two issues remain to be solved: 1) the IHN is trained on specific datasets such as the cross-modal GoogleMap dataset~\cite{zhao2021deep}, making it difficult to directly deal with RGB-T inputs well, 2) IHN focuses on the entire image region rather than the object region that the SOD task is intended to focus on.
To address the two issues, we first introduce the adaptation technique~\cite{houlsby2019parameter}, which is used to fine-tune the pre-trained model to adapt to downstream tasks in a parameter-efficient way. 
Then, we embed the semantic information that can localize object regions into the adapter to form the proposed Semantic-Adapter (S-Adapter). 

To be specific, IHN takes in a pair of source image and target image and outputs the estimated homography matrix $H$, which can map points in the target image to corresponding points in the source image. As shown in Fig.~\ref{fig::framework}, the main steps of IHN include feature extraction using a siamese encoder, correlation computation, and homography estimation.
Considering that the unaligned image pairs are mainly annotated according to the RGB modality, we take the RGB image as the source image and the thermal image as the target image to facilitate the subsequent correlation modeling.
The whole process of IHN can be formulated as:
\begin{equation}
	\begin{split}
		&H = \mathcal{H}\left( {\Psi \left( {\Phi ({I_{rgb}}),\Phi ({I_{t}})} \right)} \right),
	\end{split}
\end{equation} 
where ${\Phi (.)}$ is the IHN feature encoder, ${\Psi \left(,\right)}$ is the correlation computation function, $\mathcal{H}\left(.\right)$ refers to the homography estimator.
Then, we embed the S-Adapter into each layer of the encoder for sufficient  adaptation:
\begin{equation}
	\begin{split}
		&{\widehat F^{l}} = {F^l} + {\text{S-Adapter}_{l}}({F^l},{f_s}),
	\end{split}
\end{equation} 
where ${\widehat F^{l}}$ is the $l_{th}$ layer adapted feature in the IHN feature encoder, and will be propagated to the next layer.

\begin{figure}[t]
	\centering
	\includegraphics[width=1\linewidth]{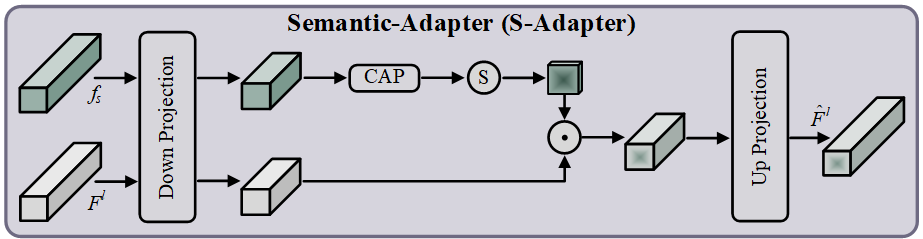}
	\caption{The details of the proposed S-Adapter.} 
	\label{fig::Adapter}
\end{figure}
\begin{figure}[t]
	\centering
	\includegraphics[width=1\linewidth]{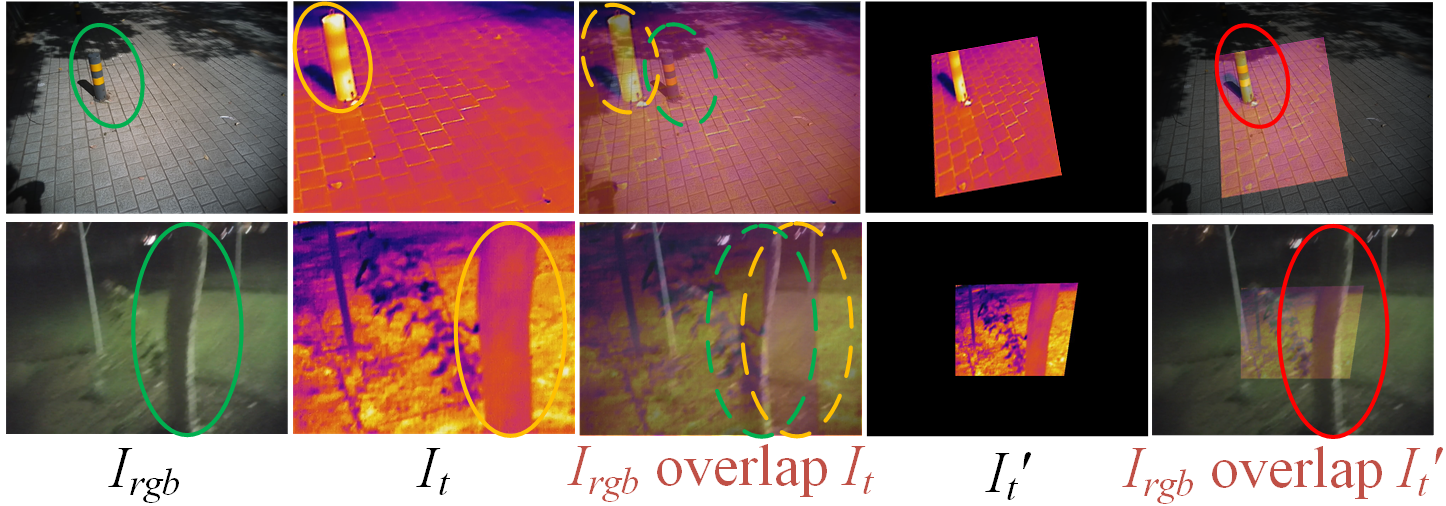}
	\caption{Examples of before and after warping.} 
	\label{fig::show}
\end{figure}
The detailed flow of S-Adapter is shown in Fig.~\ref{fig::Adapter}. 
S-Adapter is in a down-up bottleneck structure, which consists of down-projection, fusion unit $\mathcal{G}$, ReLU activation $\phi$, and up-projection sequentially. In particular, among the fusion unit, the down-projected semantic feature takes its weight distribution as the mask to guide the extracted feature to focus on object regions. The process is formulated as:
\begin{equation}
	\begin{split}
		&{\text{S-Adapter}_{l}}({F^l},{f_s}) = \phi \left( {\mathcal{G}\left( {{F^l}{\text{W}_{dn}},{f_s}{\text{W}_{dn}}} \right)} \right){\text{W}_{up}},
	\end{split}
\end{equation} 
\begin{equation}
	\begin{split}
		&\mathcal{G}(X,Y) = X \odot \sigma \left( {{\text{CAP}}\left( Y \right)} \right),
	\end{split}
\end{equation} 
where ${\text{W}_{dn}}$ and ${\text{W}_{up}}$ are the projection operation, $\odot$ is the element-wise multiplication, $\sigma$ is the Sigmoid function, and ${{\text{CAP}}\left(.\right)}$ denotes the channel-wise average pooling.

Then, the process of SHE can be formulated as follows:
\begin{equation}
	\begin{split}
		&H = \mathcal{H} \left( \Psi \left( \Phi_{\text{Adapter}}({I_{rgb}}, {f_{s}}), \Phi_{\text{Adapter}}({I_{t}}, {f_{s}}) \right) \right),
	\end{split}
\end{equation} 
where $\Phi_{\text{Adapter}}(,)$ is the encoder fine-tuned by the S-Adapter. Finally, the warped thermal image ${I'_t}$ is obtained using the estimated homography $H$:
\begin{equation}
	\begin{split}
		&{I'_t} = {\text{Warp}}({I_t},H).
	\end{split}
\end{equation}
Fig.~\ref{fig::show} shows the visual results of thermal images before and after warping for some examples, indicating that SHE can warp objects for alignment without background interference.

\begin{table*}[t!]
	\centering
	\resizebox{1\textwidth}{!}{
		\begin{tabular}{cc|ccccccccccccc||c}
			\toprule[1.5pt]
			\multicolumn{2}{c|}{Method} & MIDD$_{21}$  & CSRNet$_{21}$ & CGFNet$_{21}$ & SwinNet$_{21}$ & OSRNet$_{21}$ & TNet$_{22}$  & DCNet$_{22}$ & HRTrans$_{22}$ & MCFNet$_{23}$ & LSNet$_{23}$ & CAVER$_{23}$ & LAFB$_{24}$  & SACNet$_{24}$ & PCNet \bigstrut\\
			\hline
			\multirow{3}[2]{*}{UVT20K} & $E_m$↑ & 0.842  & 0.790  & 0.819  & 0.850  & 0.835  & 0.876  & 0.853  & 0.839  & 0.875  & 0.828  & 0.876  & 0.858  & 0.819  & \textbf{0.897} \bigstrut[t]\\
			& $S_m$↑ & 0.836  & 0.753  & 0.829  & 0.841  & 0.807  & 0.856  & 0.808  & 0.852  & 0.842  & 0.834  & 0.861  & 0.847  & 0.829  & \textbf{0.872} \\
			& $F_m$↑ & 0.743  & 0.635  & 0.709  & 0.728  & 0.730  & 0.783  & 0.776  & 0.710  & 0.800  & 0.699  & 0.790  & 0.757  & 0.709  & \textbf{0.822} \bigstrut[b]\\
			\hline
			\multirow{3}[2]{*}{UVT2000} & $E_m$↑ & 0.727  & 0.658  & 0.704  & 0.780  & 0.764  & 0.782  & 0.808  & 0.706  & 0.784  & 0.711  & 0.782  & 0.774  & 0.792  & \textbf{0.851} \bigstrut[t]\\
			& $S_m$↑ & 0.778  & 0.655  & 0.764  & 0.790  & 0.741  & 0.792  & 0.767  & 0.758  & 0.774  & 0.763  & 0.786  & 0.778  & 0.795  & \textbf{0.819} \\
			& $F_m$↑ & 0.563  & 0.420  & 0.539  & 0.592  & 0.567  & 0.610  & 0.632  & 0.525  & 0.621  & 0.527  & 0.616  & 0.594  & 0.601  & \textbf{0.686} \bigstrut[b]\\
			\hline
			\multirow{3}[2]{*}{un-VT5000} & $E_m$↑ & 0.885  & 0.713  & 0.868  & 0.901  & 0.846  & 0.905  & 0.879  & 0.899  & 0.892  & 0.892  & 0.893  & 0.908  & 0.899  & \textbf{0.936} \bigstrut[t]\\
			& $S_m$↑ & 0.830  & 0.642  & 0.833  & 0.837  & 0.800  & 0.858  & 0.812  & 0.872  & 0.836  & 0.847  & 0.850  & 0.851  & 0.872  & \textbf{0.879} \\
			& $F_m$↑ & 0.778  & 0.475  & 0.757  & 0.767  & 0.741  & 0.807  & 0.803  & 0.780  & 0.809  & 0.767  & 0.805  & 0.803  & 0.780  & \textbf{0.861} \bigstrut[b]\\
			\hline
			\multirow{3}[2]{*}{un-VT1000} & $E_m$↑ & 0.884  & 0.732  & 0.878  & 0.871  & 0.942  & 0.880  & 0.880  & 0.872  & 0.886  & 0.875  & 0.881  & 0.880  & 0.868  & \textbf{0.947} \bigstrut[t]\\
			& $S_m$↑ & 0.875  & 0.705  & 0.877  & 0.853  & 0.980  & 0.868  & 0.858  & 0.869  & 0.876  & 0.868  & 0.873  & 0.862  & 0.852  & \textbf{0.922} \\
			& $F_m$↑ & 0.837  & 0.582  & 0.817  & 0.802  & 0.951  & 0.827  & 0.850  & 0.785  & 0.850  & 0.797  & 0.838  & 0.824  & 0.803  & \textbf{0.904} \bigstrut[b]\\
			\hline
			\multirow{3}[2]{*}{un-VT821} & $E_m$↑ & 0.884  & 0.783  & 0.874  & 0.903  & 0.848  & 0.913  & 0.869  & 0.901  & 0.893  & 0.875  & 0.856  & 0.889  & 0.916  & \textbf{0.936} \bigstrut[t]\\
			& $S_m$↑ & 0.843  & 0.737  & 0.837  & 0.854  & 0.814  & 0.876  & 0.817  & 0.873  & 0.850  & 0.842  & 0.818  & 0.834  & 0.876  & \textbf{0.893} \\
			& $F_m$↑ & 0.791  & 0.619  & 0.776  & 0.783  & 0.734  & 0.829  & 0.793  & 0.782  & 0.818  & 0.754  & 0.780  & 0.791  & 0.812  & \textbf{0.869} \bigstrut[b]\\
			\hline
			\multirow{3}[2]{*}{VT5000} & $E_m$↑ & 0.897  & 0.905  & 0.922  & 0.942  & 0.908  & 0.927  & 0.920  & 0.945  & 0.924  & 0.915  & 0.924  & 0.931  & \textbf{0.957} & 0.956  \bigstrut[t]\\
			& $S_m$↑ & 0.868  & 0.868  & 0.883  & 0.912  & 0.875  & 0.895  & 0.871  & 0.912  & 0.887  & 0.877  & 0.892  & 0.893  & 0.917  & \textbf{0.920} \\
			& $F_m$↑ & 0.801  & 0.811  & 0.851  & 0.865  & 0.823  & 0.846  & 0.847  & 0.871  & 0.848  & 0.825  & 0.841  & 0.857  & \textbf{0.901} & 0.899  \bigstrut[b]\\
			\hline
			\multirow{3}[2]{*}{VT1000} & $E_m$↑ & 0.933  & 0.925  & 0.944  & 0.947  & 0.935  & 0.937  & 0.948  & 0.945  & 0.944  & 0.935  & 0.945  & 0.945  & \textbf{0.958} & \textbf{0.958} \bigstrut[t]\\
			& $S_m$↑ & 0.915  & 0.918  & 0.923  & 0.938  & 0.926  & 0.929  & 0.922  & 0.938  & 0.932  & 0.925  & 0.936  & 0.932  & 0.942  & \textbf{0.943} \\
			& $F_m$↑ & 0.882  & 0.877  & 0.906  & 0.896  & 0.892  & 0.889  & 0.911  & 0.900  & 0.902  & 0.885  & 0.903  & 0.905  & 0.923  & \textbf{0.924} \bigstrut[b]\\
			\hline
			\multirow{3}[2]{*}{VT821} & $E_m$↑ & 0.895  & 0.909  & 0.912  & 0.926  & 0.896  & 0.919  & 0.912  & 0.929  & 0.918  & 0.911  & 0.919  & 0.915  & 0.932  & \textbf{0.941} \bigstrut[t]\\
			& $S_m$↑ & 0.871  & 0.884  & 0.881  & 0.904  & 0.875  & 0.899  & 0.876  & 0.906  & 0.891  & 0.878  & 0.891  & 0.884  & 0.906  & \textbf{0.915} \\
			& $F_m$↑ & 0.804  & 0.831  & 0.845  & 0.847  & 0.814  & 0.842  & 0.841  & 0.853  & 0.844  & 0.825  & 0.839  & 0.843  & 0.868  & \textbf{0.879} \bigstrut[b]\\
			\bottomrule[1.5pt]
		\end{tabular}%
	}
	\caption{Quantitative comparisons with 13 state-of-the-art RGB-T SOD methods on two unaligned, three weakly aligned, and three aligned datasets. The best results are marked in \textbf{bold}.}
	\label{tab:compare}%
\end{table*}%
\subsection{Inter- and Intra-Modal Correlation (IIMC)}
The warped thermal image aligns the thermal modality with the corresponding and common areas in the RGB modality.
Based on this, IIMC leverages the aligned regions from the warped thermal image to model inter-modal correlations and further expands the correlations to the entire region within the RGB modality.
Considering that information loss caused by image warping hinders effective multi-modal fusion~\cite{qin2023unifusion}, we use the warped thermal image to map out the corresponding region in the RGB modality, which then interacts with the unwarped thermal modality. In addition, the semantic information is also introduced to promote the correlation modeling focusing on object regions. The process of the inter-modal correlation modeling can be formulated as:
\begin{equation}
	\begin{split}
		&f_{{\text{inter}}}^i = \mathcal{C}\left( {f_{rgb}^i \odot \mathcal{M}\left( {{{I'}_t}} \right) \odot {f_s},f_t^i} \right),
	\end{split}
\end{equation}
\begin{equation}\label{Eq::1}
	\begin{split}
		&\mathcal{C}(Q,V) = {\rm{softmax(}}\frac{{Q{K^T}}}{{\sqrt {{d_k}} }}{\rm{)}}V{\rm{ + }}Q,
	\end{split}
\end{equation} 
where $f_{{\text{inter}}}^i$ is the inter-modal correlation feature, $\mathcal{M}\left(.\right)$ is the mapping and resize operation, and $\mathcal{C}(,)$ represents the transformer-based~\cite{vaswani2017attention} correlation operation, in which $Q$, $K$, $V$ are queries, keys, and values.

Since only a subset of the region is correlated, the goal is changed to how to expand the correlated region to cover the entire region within the RGB modality. Therefore, we perform intra-modal correlation to supplement the inter-modal correlation results. Specifically, we first attach the inter-modal correlation result $f_{{\text{inter}}}^i$ to the whole RGB modality $f_{rgb}^i$ through residual connection, and then propagate the inter-modal correlation to the whole RGB modality through the correlation operation. Mathematically,
the intra-modal correlation is expressed by:
\begin{equation}\label{Eq::att}
	\begin{split}
		&f_{{\text{intra}}}^i = \mathcal{C}\left( {f_{rgb}^i + f_{{\text{inter}}}^i,f_{rgb}^i + f_{{\text{inter}}}^i} \right),
	\end{split}
\end{equation} 
where $f_{{\text{intra}}}^i$ is the output correlated feature.

Eventually, as in previous works~\cite{liu2021swinnet,wang2024alignment}, the decoder integrates the correlated features in a top-down manner to obtain final predictions, which are supervised by saliency ground-truths and optimized with a combination of binary cross-entropy loss and dice loss.

\section{Experiments}
\subsection{Experiment Setup}
\noindent\textbf{Datasets and Evaluation Metrics.}
To fully evaluate models, we conduct experiments on both aligned and unaligned datasets. 
For alignment-free models, we train our method and compared methods on the training set of UVT20K and test them on UVT2000, un-VT821, un-VT1000, and the test sets of un-VT5000 and UVT20K. Note that we include the weakly aligned datasets (i.e., un-VT821, un-VT1000, and un-VT5000) here for comprehensive evaluation.
Following previous works~\cite{tu2021multi,tu2022weakly}, we train our alignment-based model on the training set of VT5000 and test on VT821, VT1000, and the test set of VT5000.

We employ three widely used evaluation metrics to assess model performance, including enhanced-alignment ${E_m}$, structure-measure ${S_m}$, and F-measure ${F_m}$.

\noindent\textbf{Implementation Details.}
Our aligned and unaligned models are implemented on two RTX 3090 GPUs with the same settings. The AdamW optimizer with the learning rate of 1e-5, weight decay of 1e-4, batch size of 4, and training epoch of 80 is used to optimize model parameters. The input images are resized to 384 × 384 during training and inference.

\subsection{Comparison with State-of-the-Art Methods}
We compare our method with 13 state-of-the-art RGB-T SOD methods, including SACNet~\cite{wang2024alignment}, LAFB~\cite{wang2024learning}, CAVER~\cite{pang2023caver}, LSNet~\cite{zhou2023lsnet}, MCFNet~\cite{ma2023modal}, HRTransNet~\cite{tang2022hrtransnet}, DCNet~\cite{tu2022weakly}, TNet~\cite{cong2022does}, OSRNet~\cite{huo2022real}, SwinNet~\cite{liu2021swinnet}, CGFNet~\cite{wang2021cgfnet}, CSRNet~\cite{huo2021efficient}, and MIDD~\cite{tu2021multi}. To make a fair comparison, we use the results published by the authors or run their released code with default parameters. 
Table~\ref{tab:compare} presents the quantitative results, showing that our method achieves optimal performance on all eight datasets, except for a slight weakness on the $E_m$ and $F_m$ metrics of the VT5000 dataset. Compared to the sub-optimal method (i.e., CAVER), our PCNet achieves an improvement of 6.2\%, 2.8\%, and 7.8\% on the three metrics (i.e., $E_m$, $S_m$, and $F_m$) of the two unaligned datasets. It is worth noting that our method has the smallest performance gap between the weakly aligned datasets and their corresponding aligned datasets, further indicating that our method is able to effectively handle misalignment. For example, the average gap between our method on the three metrics for VT5000 and un-VT5000 is 3.8\%, while for SACNet, the gap is 9.1\%.

\begin{table}[t]
	\centering
	\resizebox{1\columnwidth}{!}{
		\begin{tabular}{c|c|ccc|ccc|ccc}
			\toprule[1pt]
			\multirow{2}[4]{*}{ID} & \multirow{2}[4]{*}{models} & \multicolumn{3}{c|}{UVT20K} & \multicolumn{3}{c|}{UVT2000} & \multicolumn{3}{c}{VT5000} \bigstrut\\
			\cline{3-11}          &       & $E_m$↑    & $S_m$↑    & $F_m$↑    & $E_m$↑    & $S_m$↑    & $F_m$↑    & $E_m$↑    & $S_m$↑    & $F_m$↑ \bigstrut\\
			\hline
			0     & PCNet & \textbf{0.897} & \textbf{0.872} & \textbf{0.822} & \textbf{0.851} & \textbf{0.819} & \textbf{0.686} & \textbf{0.956} & \textbf{0.920} & \textbf{0.899} \bigstrut\\
			\hline
			1     & w/o SHE & 0.875  & 0.849  & 0.786  & 0.819  & 0.800  & 0.660  & 0.948  & 0.915  & 0.884  \bigstrut[t]\\
			2     & w/o S-Adapter & 0.880  & 0.856  & 0.798  & 0.822  & 0.806  & 0.674  & 0.951  & 0.914  & 0.886  \\
			3     & w/ FFT & 0.890  & 0.865  & 0.810  & 0.841  & 0.813  & 0.681  & 0.953  & 0.917  & 0.889  \bigstrut[b]\\
			\hline
			4     & w/o IIMC & 0.864  & 0.843  & 0.782  & 0.817  & 0.798  & 0.656  & 0.943  & 0.907  & 0.870  \bigstrut[t]\\
			5     & w/o Intra & 0.877  & 0.851  & 0.791  & 0.831  & 0.807  & 0.668  & 0.947  & 0.909  & 0.878  \bigstrut[b]\\
			\hline
			6     & w/o Semantics & 0.888  & 0.867  & 0.804  & 0.836  & 0.810  & 0.679  & 0.954  & 0.918  & 0.894  \bigstrut\\
			\hline
			7     & w/o Thermal & 0.849  & 0.831  & 0.747  & 0.795  & 0.788  & 0.615  & 0.941  & 0.904  & 0.867  \bigstrut\\
			\bottomrule[1pt]
		\end{tabular}%
	}
	\caption{Ablation analyses of the proposed components. 'w/o': remove the component. 'FFT': full fine-tuning.}
	\label{tab:aba}%
\end{table}%

\subsection{Ablation Studies}
\noindent\textbf{Effect of SHE.}
We evaluate the impact of the SHE module by removing it (ID1 in Table~\ref{tab:aba}), which means that correlations are modeled without alignment. Compared to our full model (i.e., ID0), the average decrease is 3.2\%, 2.6\%, and 4.3\% on the three metrics (i.e., $E_m$, $S_m$, and $F_m$) across the two unaligned datasets. It is worth noting that SHE remains valid on the aligned dataset (i.e., VT5000), mainly because the fine-tuned SHE is able to adapt to different inputs and make alignments as needed.
ID2 further verifies the effectiveness of the parameter-frozen homography estimator by removing the S-Adapter. The comparison (ID2 $vs$ ID0) demonstrates the positive effect of the S-Adapter. Moreover, the weak performance of ID3 proves that full fine-tuning destroys the ability of the pre-trained homography estimator.

\noindent\textbf{Effect of IIMC.}
We replace IIMC with feature summation, which implies that the aligned multi-modal features are simply fused. Comparison with the full model (ID4 $vs$ ID0) shows that IIMC improves all metrics. In addition, ID5 shows the results of removing intra-modal correlations, indicating that the progressive modeling approach is effective.

\noindent\textbf{Effect of Semantics.}
Since the semantic information is embedded into both SHE and IIMC for guidance, we remove it to verify its effectiveness. The comparison between ID6 and ID0 shows the positive role of the semantic information.

\noindent\textbf{Effect of Thermal Modality.}
We also replace thermal inputs with corresponding visible images. This implies that our model only uses the RGB modality for prediction. The obvious performance drop in ID7 proves that the unaligned thermal modality can provide effective complementary information through correlation modeling.
In addition, the drop on the unaligned datasets is greater than that on the aligned datasets, suggesting that the RGB images in the unaligned dataset are more challenging and rely on the thermal modality as information supplement.

\begin{table}[t]
	\centering
	\resizebox{1\columnwidth}{!}{
		\begin{tabular}{c|ccc|ccc}
			\toprule[1pt]
			\multicolumn{1}{c|}{Training Set} & \multicolumn{3}{c|}{UVT20K} & \multicolumn{3}{c}{un-VT5000} \bigstrut\\
			\hline
			\multicolumn{1}{c|}{Test set} & \multicolumn{3}{c|}{UVT2000} & \multicolumn{3}{c}{UVT2000} \bigstrut\\
			\hline
			\multicolumn{1}{c|}{Metric} & $E_m$↑    & $S_m$↑    & $F_m$↑    & $E_m$↑    & $S_m$↑    & $F_m$↑ \bigstrut\\
			\hline
			PCNet (Ours) & \textbf{0.851}  & \textbf{0.819}  & \textbf{0.686}  & \textbf{0.799}  & \textbf{0.813}  & \textbf{0.646} \bigstrut[t]\\
			CAVER~\cite{pang2023caver} & 0.782  & 0.786  & 0.616  & 0.727  & 0.749  & 0.535  \\
			LSNet~\cite{zhou2023lsnet} & 0.711  & 0.763  & 0.527  & 0.679  & 0.728  & 0.478  \\
			MCFNet~\cite{ma2023modal} & 0.784  & 0.774  & 0.621  & 0.727  & 0.739  & 0.535  \bigstrut[b]\\
			\bottomrule[1pt]
		\end{tabular}%
	}
	\caption{Comparison of models using different training sets.}
	\label{tab:datasetEff}%
\end{table}%
\noindent\textbf{Effect of UVT20K.}
We further analyze the effectiveness of the proposed UVT20K dataset by comparing the model performance of using the other training set. As in work~\cite{wang2024alignment}, we train our model and some recent advanced models using the training set of un-VT5000. For a fair comparison, we only evaluate them on the UVT2000 dataset instead of the test sets of UVT20K or un-VT5000 with the same distribution. The results in Table~\ref{tab:datasetEff} show that the model trained with the large-scale UVT20K has significant performance advantages, indicating that UVT20K contributes to the research of alignment-free RGB-T SOD. Moreover, the optimal performance on both training sets demonstrates the generalization of our method.

\section{Conclusion}
In this paper, we construct a large-scale dataset UVT20K for alignment-free RGB-T SOD. To the best of our knowledge, UVT20K is the largest multi-modal SOD dataset, containing 20,000 samples divided into training and test sets. It covers a wide range of scenes and object categories, offers a comprehensive set of annotations, and presents significant challenges.
With these features, the UVT20K will make significant contributions to research on RGB-T SOD and unaligned image pairs. In addition, we propose a Progressive Correlation Network (PCNet) that aligns the corresponding regions and successively models inter-modal and intra-modal correlations for accurate saliency predictions in unaligned image pairs.
Extensive experiments demonstrate the potential of the proposed UVT20K dataset and the effectiveness of our method.
In future work, we will explore more challenges in unaligned image pairs, such as varying camera viewpoints and inconsistent image depth of field.

\section{Acknowledgments}
This work was in part supported by the University Synergy Innovation Program of Anhui Province (No. GXXT-2022-014), the National Natural Science Foundation of China (No. 62376004 and No. 62376005), and the Natural Science Foundation of Anhui Province (NO. 2208085J18).

\bibliography{aaai25}

\begin{thebibliography}{33}
\providecommand{\natexlab}[1]{#1}

\bibitem[{Cao et~al.(2022)Cao, Hu, Sheng, and Shen}]{cao2022iterative}
Cao, S.-Y.; Hu, J.; Sheng, Z.; and Shen, H.-L. 2022.
\newblock Iterative deep homography estimation.
\newblock In \emph{Proceedings of the IEEE/CVF conference on computer vision
  and pattern recognition}, 1879--1888.

\bibitem[{Cong et~al.(2022)Cong, Zhang, Zhang, Zheng, Zhao, Huang, and
  Kwong}]{cong2022does}
Cong, R.; Zhang, K.; Zhang, C.; Zheng, F.; Zhao, Y.; Huang, Q.; and Kwong, S.
  2022.
\newblock Does thermal really always matter for RGB-T salient object detection?
\newblock \emph{IEEE Transactions on Multimedia}, 25: 6971--6982.

\bibitem[{Fan et~al.(2019)Fan, Wang, Cheng, and Shen}]{fan2019video}
Fan, D.; Wang, W.; Cheng, M.; and Shen, J. 2019.
\newblock Shifting More Attention to Video Salient Object Detection.
\newblock In \emph{{IEEE} Conference on Computer Vision and Pattern
  Recognition}, 8554--8564.

\bibitem[{Fan et~al.(2020)Fan, Lin, Zhang, Zhu, and Cheng}]{fan2020rethinking}
Fan, D.-P.; Lin, Z.; Zhang, Z.; Zhu, M.; and Cheng, M.-M. 2020.
\newblock Rethinking RGB-D salient object detection: Models, data sets, and
  large-scale benchmarks.
\newblock \emph{IEEE Transactions on neural networks and learning systems},
  32(5): 2075--2089.

\bibitem[{Houlsby et~al.(2019)Houlsby, Giurgiu, Jastrzebski, Morrone,
  De~Laroussilhe, Gesmundo, Attariyan, and Gelly}]{houlsby2019parameter}
Houlsby, N.; Giurgiu, A.; Jastrzebski, S.; Morrone, B.; De~Laroussilhe, Q.;
  Gesmundo, A.; Attariyan, M.; and Gelly, S. 2019.
\newblock Parameter-efficient transfer learning for NLP.
\newblock In \emph{International Conference on Machine Learning}, 2790--2799.
  PMLR.

\bibitem[{Huo et~al.(2021)Huo, Zhu, Zhang, Liu, and Shu}]{huo2021efficient}
Huo, F.; Zhu, X.; Zhang, L.; Liu, Q.; and Shu, Y. 2021.
\newblock Efficient context-guided stacked refinement network for RGB-T salient
  object detection.
\newblock \emph{IEEE Transactions on Circuits and Systems for Video
  Technology}, 32(5): 3111--3124.

\bibitem[{Huo et~al.(2022)Huo, Zhu, Zhang, Liu, and Yu}]{huo2022real}
Huo, F.; Zhu, X.; Zhang, Q.; Liu, Z.; and Yu, W. 2022.
\newblock Real-time one-stream semantic-guided refinement network for
  RGB-thermal salient object detection.
\newblock \emph{IEEE Transactions on Instrumentation and Measurement}, 71:
  1--12.

\bibitem[{Li et~al.(2017)Li, Xu, Ren, and Wang}]{li2017closed}
Li, S.; Xu, M.; Ren, Y.; and Wang, Z. 2017.
\newblock Closed-form optimization on saliency-guided image compression for
  HEVC-MSP.
\newblock \emph{IEEE Transactions on Multimedia}, 20(1): 155--170.

\bibitem[{Liu et~al.(2021{\natexlab{a}})Liu, Zhang, Shao, and
  Han}]{liu2021learning}
Liu, N.; Zhang, N.; Shao, L.; and Han, J. 2021{\natexlab{a}}.
\newblock Learning selective mutual attention and contrast for RGB-D saliency
  detection.
\newblock \emph{IEEE Transactions on Pattern Analysis and Machine
  Intelligence}, 44(12): 9026--9042.

\bibitem[{Liu et~al.(2021{\natexlab{b}})Liu, Lin, Cao, Hu, Wei, Zhang, Lin, and
  Guo}]{liu2021swin}
Liu, Z.; Lin, Y.; Cao, Y.; Hu, H.; Wei, Y.; Zhang, Z.; Lin, S.; and Guo, B.
  2021{\natexlab{b}}.
\newblock Swin transformer: Hierarchical vision transformer using shifted
  windows.
\newblock In \emph{Proceedings of the IEEE/CVF international conference on
  computer vision}, 10012--10022.

\bibitem[{Liu et~al.(2021{\natexlab{c}})Liu, Tan, He, and
  Xiao}]{liu2021swinnet}
Liu, Z.; Tan, Y.; He, Q.; and Xiao, Y. 2021{\natexlab{c}}.
\newblock SwinNet: Swin transformer drives edge-aware RGB-D and RGB-T salient
  object detection.
\newblock \emph{IEEE Transactions on Circuits and Systems for Video
  Technology}, 32(7): 4486--4497.

\bibitem[{Lu et~al.(2021)Lu, Li, Yan, Tang, and Luo}]{lu2021rgbt}
Lu, A.; Li, C.; Yan, Y.; Tang, J.; and Luo, B. 2021.
\newblock RGBT tracking via multi-adapter network with hierarchical divergence
  loss.
\newblock \emph{IEEE Transactions on Image Processing}, 30: 5613--5625.

\bibitem[{Ma et~al.(2023)Ma, Song, Dong, Tian, and Yan}]{ma2023modal}
Ma, S.; Song, K.; Dong, H.; Tian, H.; and Yan, Y. 2023.
\newblock Modal complementary fusion network for RGB-T salient object
  detection.
\newblock \emph{Applied Intelligence}, 53(8): 9038--9055.

\bibitem[{Pang et~al.(2023)Pang, Zhao, Zhang, and Lu}]{pang2023caver}
Pang, Y.; Zhao, X.; Zhang, L.; and Lu, H. 2023.
\newblock {CAVER:} Cross-Modal View-Mixed Transformer for Bi-Modal Salient
  Object Detection.
\newblock \emph{{IEEE} Trans. Image Process.}, 32: 892--904.

\bibitem[{Piao et~al.(2019)Piao, Ji, Li, Zhang, and Lu}]{piao2019depth}
Piao, Y.; Ji, W.; Li, J.; Zhang, M.; and Lu, H. 2019.
\newblock Depth-induced multi-scale recurrent attention network for saliency
  detection.
\newblock In \emph{Proceedings of the IEEE/CVF international conference on
  computer vision}, 7254--7263.

\bibitem[{Qin et~al.(2023)Qin, Chen, Chen, Chen, and Li}]{qin2023unifusion}
Qin, Z.; Chen, J.; Chen, C.; Chen, X.; and Li, X. 2023.
\newblock Unifusion: Unified multi-view fusion transformer for spatial-temporal
  representation in bird's-eye-view.
\newblock In \emph{Proceedings of the IEEE/CVF International Conference on
  Computer Vision}, 8690--8699.

\bibitem[{Tang et~al.(2022)Tang, Liu, Tan, and He}]{tang2022hrtransnet}
Tang, B.; Liu, Z.; Tan, Y.; and He, Q. 2022.
\newblock HRTransNet: HRFormer-driven two-modality salient object detection.
\newblock \emph{IEEE Transactions on Circuits and Systems for Video
  Technology}, 33(2): 728--742.

\bibitem[{Tang et~al.(2019)Tang, Fan, Wang, Tu, and Li}]{tang2019rgbt}
Tang, J.; Fan, D.; Wang, X.; Tu, Z.; and Li, C. 2019.
\newblock RGBT salient object detection: Benchmark and a novel cooperative
  ranking approach.
\newblock \emph{IEEE Transactions on Circuits and Systems for Video
  Technology}, 30(12): 4421--4433.

\bibitem[{Tu et~al.(2021)Tu, Li, Li, Lang, and Tang}]{tu2021multi}
Tu, Z.; Li, Z.; Li, C.; Lang, Y.; and Tang, J. 2021.
\newblock Multi-interactive dual-decoder for RGB-thermal salient object
  detection.
\newblock \emph{IEEE Transactions on Image Processing}, 30: 5678--5691.

\bibitem[{Tu et~al.(2022)Tu, Li, Li, and Tang}]{tu2022weakly}
Tu, Z.; Li, Z.; Li, C.; and Tang, J. 2022.
\newblock Weakly alignment-free RGBT salient object detection with deep
  correlation network.
\newblock \emph{IEEE Transactions on Image Processing}, 31: 3752--3764.

\bibitem[{Tu et~al.(2020)Tu, Ma, Li, Li, Xu, and Liu}]{tu2020rgbt}
Tu, Z.; Ma, Y.; Li, Z.; Li, C.; Xu, J.; and Liu, Y. 2020.
\newblock {RGBT} Salient Object Detection: {A} Large-scale Dataset and
  Benchmark.
\newblock \emph{CoRR}, abs/2007.03262.

\bibitem[{Tu et~al.(2019)Tu, Xia, Li, Wang, Ma, and Tang}]{tu2019rgb}
Tu, Z.; Xia, T.; Li, C.; Wang, X.; Ma, Y.; and Tang, J. 2019.
\newblock RGB-T image saliency detection via collaborative graph learning.
\newblock \emph{IEEE Transactions on Multimedia}, 22(1): 160--173.

\bibitem[{Vaswani(2017)}]{vaswani2017attention}
Vaswani, A. 2017.
\newblock Attention is All You Need.
\newblock \emph{arXiv preprint arXiv:1706.03762}.

\bibitem[{Wang et~al.(2018)Wang, Li, Ma, Zheng, Tang, and Luo}]{wang2018rgb}
Wang, G.; Li, C.; Ma, Y.; Zheng, A.; Tang, J.; and Luo, B. 2018.
\newblock RGB-T saliency detection benchmark: Dataset, baselines, analysis and
  a novel approach.
\newblock In \emph{Image and Graphics Technologies and Applications: 13th
  Conference on Image and Graphics Technologies and Applications, IGTA 2018,
  Beijing, China, April 8--10, 2018, Revised Selected Papers 13}, 359--369.
  Springer.

\bibitem[{Wang et~al.(2021)Wang, Song, Bao, Huang, and Yan}]{wang2021cgfnet}
Wang, J.; Song, K.; Bao, Y.; Huang, L.; and Yan, Y. 2021.
\newblock CGFNet: Cross-guided fusion network for RGB-T salient object
  detection.
\newblock \emph{IEEE Transactions on Circuits and Systems for Video
  Technology}, 32(5): 2949--2961.

\bibitem[{Wang et~al.(2024{\natexlab{a}})Wang, Lin, Li, Tu, and
  Luo}]{wang2024alignment}
Wang, K.; Lin, D.; Li, C.; Tu, Z.; and Luo, B. 2024{\natexlab{a}}.
\newblock Alignment-Free RGBT Salient Object Detection: Semantics-guided
  Asymmetric Correlation Network and A Unified Benchmark.
\newblock \emph{IEEE Transactions on Multimedia}.

\bibitem[{Wang et~al.(2024{\natexlab{b}})Wang, Tu, Li, Zhang, and
  Luo}]{wang2024learning}
Wang, K.; Tu, Z.; Li, C.; Zhang, C.; and Luo, B. 2024{\natexlab{b}}.
\newblock Learning Adaptive Fusion Bank for Multi-modal Salient Object
  Detection.
\newblock \emph{IEEE Transactions on Circuits and Systems for Video
  Technology}.

\bibitem[{Wang et~al.(2022)Wang, Lai, Fu, Shen, Ling, and
  Yang}]{wang2022survey}
Wang, W.; Lai, Q.; Fu, H.; Shen, J.; Ling, H.; and Yang, R. 2022.
\newblock Salient Object Detection in the Deep Learning Era: An In-Depth
  Survey.
\newblock \emph{{IEEE} Trans. Pattern Anal. Mach. Intell.}, 44(6): 3239--3259.

\bibitem[{Yuan et~al.(2024)Yuan, Shi, Wang, Wang, and Wei}]{yuan2024improving}
Yuan, M.; Shi, X.; Wang, N.; Wang, Y.; and Wei, X. 2024.
\newblock Improving RGB-infrared object detection with cascade alignment-guided
  transformer.
\newblock \emph{Information Fusion}, 105: 102246.

\bibitem[{Zhang et~al.(2020)Zhang, Liu, Wang, Lei, Wang, and
  Lu}]{zhang2020track}
Zhang, P.; Liu, W.; Wang, D.; Lei, Y.; Wang, H.; and Lu, H. 2020.
\newblock Non-rigid object tracking via deep multi-scale spatial-temporal
  discriminative saliency maps.
\newblock \emph{Pattern Recognit.}, 100: 107130.

\bibitem[{Zhao, Huang, and Zhang(2021)}]{zhao2021deep}
Zhao, Y.; Huang, X.; and Zhang, Z. 2021.
\newblock Deep lucas-kanade homography for multimodal image alignment.
\newblock In \emph{Proceedings of the IEEE/CVF conference on computer vision
  and pattern recognition}, 15950--15959.

\bibitem[{Zhou et~al.(2023)Zhou, Zhu, Lei, Yang, and Yu}]{zhou2023lsnet}
Zhou, W.; Zhu, Y.; Lei, J.; Yang, R.; and Yu, L. 2023.
\newblock LSNet: Lightweight spatial boosting network for detecting salient
  objects in RGB-thermal images.
\newblock \emph{IEEE Transactions on Image Processing}, 32: 1329--1340.

\bibitem[{Zhu et~al.(2021)Zhu, Li, Tang, Luo, and Wang}]{zhu2021rgbt}
Zhu, Y.; Li, C.; Tang, J.; Luo, B.; and Wang, L. 2021.
\newblock RGBT tracking by trident fusion network.
\newblock \emph{IEEE Transactions on Circuits and Systems for Video
  Technology}, 32(2): 579--592.

\end{thebibliography}

\end{document}